\documentclass[lettersize,journal]{IEEEtran}
\usepackage{amsmath,amsfonts}
\usepackage{array}
\usepackage{color}
\usepackage{textcomp}
\usepackage{stfloats}
\usepackage{url}
\usepackage{graphicx}
\usepackage{subcaption}
\usepackage{cite}
\usepackage{diagbox} 
\usepackage{booktabs}  
\usepackage{multirow}  
\usepackage{makecell}  
\usepackage{cite}
\usepackage{url} 
\begin{document}

\title{Robust Second-order LiDAR Bundle Adjustment Algorithm Using Mean Squared Group Metric}

\author{Tingchen Ma, Yongsheng Ou* and Sheng Xu*
\thanks{This work was supported by the National Natural Science Foundation of China (Grants No. 62173319, 62063006)}
\thanks{* Corresponding author, e-mail: ys.ou@siat.ac.cn.}
\thanks{Tingchen Ma is with the Shenzhen Institute of Advanced Technology, Chinese Academy of Sciences, Shenzhen 518055, China, and also with the Shenzhen College of Advanced Technology, University of Chinese Academy of Sciences, Shenzhen 518055, China.}
\thanks{Yongsheng Ou is with the Guangdong Provincial Key Lab of Robotics and Intelligent System, Shenzhen Institute of Advanced Technology, Chinese Academy of Sciences, Shenzhen 518055, China}
}

\markboth{Journal of \LaTeX\ Class Files,~Vol.~14, No.~8, August~2021}%
{Shell \MakeLowercase{\textit{et al.}}: A Sample Article Using IEEEtran.cls for IEEE Journals}


\maketitle
\pagestyle{empty}  
\thispagestyle{empty} 

\begin{abstract}
The bundle adjustment (BA) algorithm is a widely used nonlinear optimization technique in the backend of Simultaneous Localization and Mapping (SLAM) systems. By leveraging the co-view relationships of landmarks from multiple perspectives, the BA method constructs a joint estimation model for both poses and landmarks, enabling the system to generate refined maps and reduce front-end localization errors. However, there are unique challenges when applying the BA for LiDAR data, due to the large volume of 3D points. Exploring a robust LiDAR BA estimator and achieving accurate solutions is a very important issue. In this work, firstly we propose a novel mean square group metric (MSGM) to build the optimization objective in the LiDAR BA algorithm. This metric applies mean square transformation to uniformly process the measurement of plane landmarks from one sampling period. The transformed metric ensures scale interpretability, and does not requie a time-consuming point-by-point calculation. Secondly, by integrating a robust kernel function, the metrics involved in the BA algorithm are reweighted, and thus enhancing the robustness of the solution process. Thirdly, based on the proposed robust LiDAR BA model, we derived an explicit second-order estimator (RSO-BA). This estimator employs analytical formulas for Hessian and gradient calculations, ensuring the precision of the BA solution. Finally, we verify the merits of the proposed RSO-BA estimator against existing implicit second-order and explicit approximate second-order estimators using the publicly available datasets. The experimental results demonstrate that the RSO-BA estimator outperforms its counterparts regarding registration accuracy and robustness, particularly in large-scale or complex unstructured environments.
\end{abstract}


\section{Introduction}

Light detection and ranging (LiDAR) is a three-dimensional scanning sensor with the continuous-time sampling characteristic. It is widely used in geographic surveying, urban monitoring, and mobile robots. In mobile robots, LiDAR-based Simultaneous Localization and Mapping (SLAM) systems are crucial for constructing accurate 3D maps of the environment and ensuring robust positioning. Currently, LiDAR SLAM systems are categorized into two main approaches, i.e., direct methods and feature-based methods.

LiDAR SLAM systems based on direct methods \cite{xu2022fast, dellenbach2022ct}  utilize grid downsampling to process measurement point clouds. By reducing the resolution of the point clouds, these systems require less computation, enabling real-time positioning and mapping. Specifically, each current LiDAR frame undergoes a scan-to-model registration process for pose estimation. Then, the point cloud will be transformed into the map coordinate system and used to update the point-based map (relative to the feature-based map). In the above scheme, the minimum unit of the map model is a single point in the 3D space. However, due to the sparsity of sampled LiDAR data, it is challenging to consistently hit the same spatial position across consecutive frames. As a result, point-based bundle adjustment (BA) algorithms are not applicable in such systems. On the other hand, feature-based LiDAR SLAM systems employ techniques like scanline smoothness \cite{zhang2014loam, lin2020loam}, voxel segmentation \cite{liu2021balm, liu2023efficient} and region growing \cite{zhou2020efficient, zhou2021lidar} to extract structured edge and surface feature point sets from measurement point clouds. These extracted features are then used to accomplish localization and mapping tasks. Compared to 3D points, stable data correlation between frames is more easily obtained for structural feature landmarks. LiDAR BA technology based on structural features has gotten significant advancements in recent years.

Some studies \cite{geneva2018lips, zhou2020efficient, zhou2021lidar} have designed many LiDAR BA models similar to visual BA \cite{triggs2000bundle} (referred to as explicit BA in this paper). During state estimation, all participating landmarks and robot poses are updated simultaneously, with the Schur complement technique often employed to accelerate the solution of the explicit BA model. In contrast, other approaches \cite{liu2021balm, liu2023efficient}  decompose the standard BA problem into two steps: eigenvalue fitting and multi-view pose registration (defined here as implicit BA). Unlike explicit BA schemes, \cite{liu2021balm, liu2023efficient} derived a second-order estimator using the analytical Hessian matrix and gradient vector, resulting in higher estimation accuracy. In the latest developments on two types of schemes \cite{zhou2021lidar} and  \cite{liu2023efficient}, the concept of "point clustering" has been introduced to avoid point-based operations during the solving process of linear systems, thereby improving computational efficiency. The error metric constructed based on the point clustering matrix equals the sum of error metrics corresponding to all measurement points that make up the matrix. Consequently, these schemes treat all landmarks equally during the BA-solving process. However, in practical scenarios, applying appropriate weighting to the metric (using robust kernel functions \cite{zach2014robust}) can significantly improve the system's accuracy and adaptability in complex environments. Additionally, in the two-step solution method proposed by implicit BA, eigenvalue fitting depends on the initial value (provided by the front end) or the pose estimation result from the previous iteration, which may negatively impact the robustness and accuracy of the solution.

In response to the above challenges, we propose a novel estimator to achieve high-precision map refinement in the backend of LiDAR SLAM systems. The key contributions of this article are summarized below.

\begin{itemize}
\item[$\bullet$] {We propose a new MSGM that considers the number of measurement points during its construction, ensuring the interpretability of scale. This metric allows for developing a robust LiDAR BA model by incorporating robust kernel functions.}
\end{itemize}

\begin{itemize}
\item[$\bullet$] {We derived the analytical Hessian matrix and gradient vector required by the estimator, and then, the new RSO-BA estimator is developed. The RSO-BA method is utilized by the LiDAR SLAM system with experimental verifications.}
\end{itemize}

\section{Related work}

\subsection{Point Cloud Registration}

Point cloud registration is a fundamental technique in the field of 3D vision. It aims to estimate the 3/6 degree of freedom pose transformation between the current frame point cloud and the target model (whether another point cloud or a map) through data association. The ICP \cite{besl1992method} algorithm employs nearest neighbor search to update data associations iteratively, refining the state estimation until the error converges. While the ICP algorithm is highly generalizable, it requires a certain level of point cloud density in both the frame and the model. Considering the universality of surface features, Chen et al. \cite{chen1992object}  proposed a point-to-plane registration algorithm. This approach derives surface features by fitting the local point set in the target model using principal component analysis (PCA) \cite{hotelling1933analysis}. The G-ICP \cite{segal2009generalized} algorithm considers using the current and target point covariance distribution during registration, enabling robust performance even in unstructured environments. Further enhancements, such as the VG-ICP \cite{koide2021voxelized} and N-ICP \cite{serafin2015nicp} algorithms, introduce multi-distribution modeling and normal vector constraints, respectively, thereby improving registration accuracy.

As a commonly used 3D scanning sensor, Lidar has the characteristic of continuous time sampling. When the LiDAR is in motion, distortion is inevitably introduced to the measurement point cloud. Directly applying traditional registration algorithms to such distorted data can result in significant state estimation errors. To address this issue, Zhang et al. \cite{zhang2014loam} proposed assuming uniform motion of the robot during the point cloud frame sampling process. In scan-to-scan registration, linear interpolation compensates for motion distortion in the point cloud. Dellenbach et al. \cite{dellenbach2022ct} proposed a CT-ICP algorithm that simultaneously estimates the beginning and end states of a point cloud frame. This algorithm uses linear and spherical interpolation methods to handle translation and quaternion components. Point clouds containing distortions can be directly registered to the map. 

As a high-frequency internal sensor, the Inertial Measurement Unit (IMU) can complement LiDAR well \cite{shan2020lio}. Frameworks like \cite{xu2022fast} utilize IMU-based bidirectional propagation algorithms to preprocess point cloud distortions. The IEKF \cite{bell1993iterated} algorithm is then employed to complete state estimation. Typically, undistorted LiDAR point clouds and single-frame pose estimation can be used as initial values for the Bundle Adjustment (BA) problem. In the LiDAR SLAM backend, the BA algorithm, which considers the co-view constraint of landmarks, is used to generate higher-precision map models.

\subsection{Bundle Adjustment}

The bundle adjustment \cite{triggs2000bundle} algorithm is widely used in 3D reconstruction and SLAM. It leverages the co-view constraints of landmarks to construct a joint estimation model for both landmarks and multi-frame poses, allowing for the refinement of the local map. Initially, BA algorithms were primarily used in 3D vision, where spatial points, edges, and planes are re-projected onto 2D images to obtain error metrics for the BA model. To address mismatches, kernel functions such as Huber \cite{beran1977robust}, and Cauchy \cite{black1996unification} are employed to reweight the error metrics, resulting in a more robust BA model. This robust BA technology has been integrated into many open-source visual SLAM \cite{qin2018vins, leutenegger2013keyframe} systems. On the other hand, the standard BA model often requires simultaneous estimation of thousands of landmark parameters, which can be computationally intensive. To expedite the solution of the normal equations, the Schur complement technique is frequently used, as seen in the open-source library like g2o \cite{kummerle2011g}.

Unlike visual BA algorithms, the LiDAR BA algorithms face more challenges due to the unique sampling method of the sensor. Geneva et al. \cite{geneva2018lips} proposed a plane-to-plane error metric to construct the BA model. Before solving the BA model, the plane landmark measurements are obtained by processing the local point cloud in the LiDAR coordinate system using the PCA fitting algorithm. The plane-to-plane error metric avoids point-by-point calculations during BA solving, leading to high computational efficiency. Furthermore, Zhou et al. \cite{zhou2020efficient} conducted a theoretical analysis comparing point-to-plane and plane-to-plane error metrics, with experimental results confirming that point-to-plane error metrics offer higher estimation accuracy. In related works \cite{liu2023efficient, zhou2021lidar}, methods based on "point clustering" have been proposed to process all measurement points for landmarks from each perspective. This approach makes the time complexity of the BA problem dependent only on the size of the sliding window and the number of landmarks involved in optimization. The BA problem is solved using an approximate second-order method based on Gauss-Newton.

Considering the specific structure of LiDAR BA models, which are based on edge and surface landmarks, some studies \cite{liu2021balm, liu2023efficient} have proposed efficient two-step iterative solution schemes. These schemes sequentially perform eigenvalue fitting and multi-view registration to refine the map. Unlike the explicit BA scheme, the model dimensions of implicit BA are only related to the number of keyframes in the sliding window. However, the eigenvalue fitting step theoretically overlooks the impact of pose updates during each iteration, potentially leading to a loss of estimation accuracy. Additionally, while "point clustering" can enhance the solution efficiency of the BA model, it sacrifices scale interpretability in the corresponding error metric, making robust BA techniques unusable and introducing potential risks in estimation. To address these issues, we proposed a novel mean square group metric that maintains scale interpretability, allowing for the construction of robust BA models.  We also derived an explicit second-order estimator for solving the robust BA model. The proposed estimator performs robustly and accurately in complex scenarios. 

\section{Method}

\subsection{Problem Formulation}

Assume the map coordinate system is denoted as $m$, and the measurement point cloud obtained by the LiDAR is defined in the LiDAR coordinate system $l$. The backend of the LiDAR SLAM system maintains a sliding window containing $N_w$ keyframes. The pose of keyframe ${\mathbb{F}_i}$ in the map coordinate system is represented as ${{\bf{T}}_i} \in {\rm{SE}}(3)$. The set of poses within the sliding window can be denoted as ${\bf{\chi }} = \left\{ {{{\bf{T}}_i}} \right\}_{i = 1}^{{N_w}}$.

The set of plane landmark (the most common feature in the environment) observed by keyframe ${\mathbb{F}_i}$ is denoted as ${{\bf{\Gamma }}_i} = \left\{ {{{\bf{\pi }}_j}} \right\}_{j = 1}^{{N_i}}$. The 3D measurement point set corresponding to each landmark ${{\bf{\pi }}_j}$ is represented as ${{\bf{{\rm Z}}}_{i,j}} = \left\{ {{\bf{p}}_{i,j,k}^l} \right\}_{k = 1}^{{N_{i,j}}}$. Each plane landmark maintains a fixed set ${{\bf{F}}_j} = \left\{ {{\bf{p}}_{j,o}^m} \right\}_{o = 1}^{N_j^f}$ in the map coordinate system. The points in ${{\bf{F}}_j}$ are obtained by transforming local measurements to the map coordinate system when a keyframe exits the sliding window. Assuming ${\bf{\tilde p}}_{_{i,j,k}}^l$ and ${\bf{\tilde p}}_{_{j,o}}^m$ are homogeneous coordinates of ${\bf{p}}_{i,j,k}^l$ and ${\bf{p}}_{j,o}^m$, respectively, we can define the distance ${e_{i,j,k}}$ from a local measurement point ${\bf{\tilde p}}_{_{i,j,k}}^l$ to the plane landmark ${{\bf{\pi }}_j}$ 

\begin{equation}
{e_{i,j,k}} = {\bf{\pi }}_j^T{{\bf{T}}_i}{\bf{\tilde p}}_{_{i,j,k}}^l = {\bf{\tilde p}}_{_{i,j,k}}^{l{\kern 1pt} {\kern 1pt} {\kern 1pt} {\kern 1pt} T}{\bf{T}}_i^T{{\bf{\pi }}_j}. 
\end{equation}

The distance from a fixed map point to the plane landmark is given by

\begin{equation}
e_{j,o}^f = {\bf{\pi }}_j^T{\bf{\tilde p}}_{_{j,o}}^m = {\bf{\tilde p}}_{_{j,o}}^{m{\kern 1pt} {\kern 1pt} {\kern 1pt} {\kern 1pt} T}{{\bf{\pi }}_j},
\end{equation}

\noindent where, the plane landmark is represented by ${{\bf{\pi }}_j} = {\left[ {{{\bf{n}}_j},{d_j}} \right]^T}$, with ${{\bf{n}}_j} = {\left[ {{n_{j,x}},{n_{j,y}},{n_{j,z}}} \right]^T}$ as normal vector, $d_j$ as the distance from the origin to the plane.  Assuming that set ${\bf{\Pi }} = \left\{ {{{\bf{\pi }}_q}} \right\}_{q = 1}^{N_w^{\bf{\pi }}}$ contains all effective plane landmarks within the sliding window. The standard BA model can be denoted as \cite{triggs2000bundle}

\begin{equation}
C = \arg \mathop {\min }\limits_{\bf{X}} \sum\limits_{i = 1}^{{N_w}} {\sum\limits_{j = 1}^{{N_i}} {\sum\limits_{k = 1}^{{N_{i,j}}} {\left\| {{e_{i,j,k}}} \right\|_2^2} } }  + \sum\limits_{j = 1}^{{N_i}} {\sum\limits_{o = 1}^{N_j^f} {\left\| {e_{j,o}^f} \right\|_2^2} } , 
\end{equation}

\noindent where, ${\bf{X}} = \left\{ {{\bf{\chi }},{\bf{\Pi }}} \right\}$ represent all the optimized variables in (3). In Section III.B, we introduce a robust LiDAR BA model considering the kernel function. Section III.C further derives an explicit second-order estimator for solving this BA model. The proposed estimator can be integrated into the system shown in Figure 1 and accomplish the map refinement task.

\begin{figure}[t]
\centering
\includegraphics[scale=0.56]{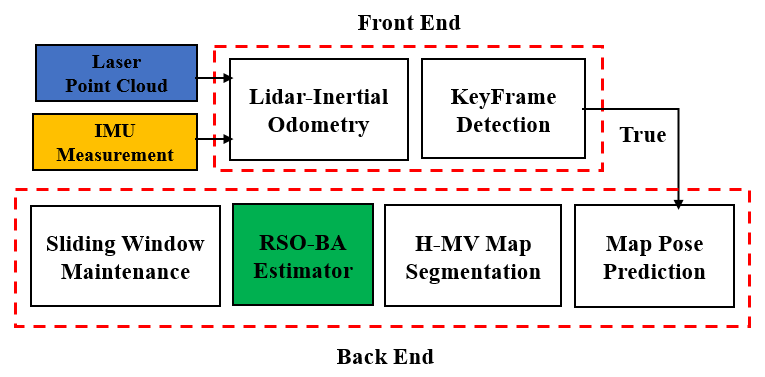}
\caption{System flowchart. The green box represents the proposed explicit second-order BA estimator.}
\label{fig_framework}
\end{figure}

\subsection{Mean Square Group Metric and Robust LiDAR BA Model}

Given the local measurement point set ${{\bf{{\rm Z}}}_{i,j}} = \left\{ {{\bf{p}}_{i,j,k}^l} \right\}_{k = 1}^{{N_{i,j}}}$ and the fixed point set ${{\bf{F}}_j} = \left\{ {{\bf{p}}_{j,o}^m} \right\}_{o = 1}^{N_j^f}$ of landmark ${{\bf{\pi }}_j}$ at keyframe ${\mathbb{F}_i}$, we can organize ${{\bf{{\rm Z}}}_{i,j}}$ into the column vector ${{\bf{P}}_{i,j}} = {\left[ {...,{\bf{\tilde p}}_{i,j,k}^l,...} \right]^T}$ and ${{\bf{F}}_j}$ into the column vector ${{\bf{P}}_j} = {\left[ {...,{\bf{\tilde p}}_{j,o}^m,...} \right]^T}$. Consequently, (3) can be rewritten as

\begin{equation}
C = \arg \mathop {\min }\limits_{\bf{X}} \sum\limits_{i = 1}^{{N_w}} {\sum\limits_{j = 1}^{{N_i}} {{c_{i,j}}} }  + \sum\limits_{j = 1}^{{N_i}} {{c_j}}, 
\end{equation}

\noindent where, ${c_{i,j}} = {\bf{\pi }}_j^T{{\bf{T}}_i}{{\bf{Q}}_{i,j}}{\bf{T}}_i^T{{\bf{\pi }}_j}$ represents the integrated group metric and ${c_j} = {\bf{\pi }}_j^T{{\bf{M}}_j}{{\bf{\pi }}_j}$ denotes the fixed group metric. The integrated group matrix ${{\bf{Q}}_{i,j}}$ and fixed group matrix ${{\bf{M}}_j}$ are defined as follows
\begin{equation}
{{\bf{Q}}_{i,j}} = {\bf{P}}_{i,j}^T{{\bf{P}}_{i,j}} 
\end{equation}
\begin{equation}
{{\bf{M}}_j} = {\bf{P}}_j^T{{\bf{P}}_j}.
\end{equation}

For the standard BA model corresponding to (3), we can enhance the robustness of the estimation model by applying a robust kernel function to each point-to-plane error metric. However, the error metrics ${c_{i,j}}$ and ${c_j}$ in (4) compromise scale interpretability due to the use of the group matrix, making it difficult to determine the appropriate threshold for the robust kernel function. Therefore, we introduce the number of measurement points to adjust ${c_{i,j}}$ and ${c_j}$. The robust BA model becomes

\begin{equation}
C' = \arg \mathop {\min }\limits_{\bf{X}} \sum\limits_{i = 1}^{{N_w}} {\sum\limits_{j = 1}^{{N_i}} {\rho \left( {{{{c_{i,j}}} \mathord{\left/
 {\vphantom {{{c_{i,j}}} {{N_{i,j}}}}} \right.
 \kern-\nulldelimiterspace} {{N_{i,j}}}}} \right)} }  + \sum\limits_{j = 1}^{{N_i}} {\rho \left( {{{{c_j}} \mathord{\left/
 {\vphantom {{{c_j}} {N_j^f}}} \right.
 \kern-\nulldelimiterspace} {N_j^f}}} \right)}, 
\end{equation}

\noindent where, $\rho \left(  \cdot  \right)$ represents the robust kernel function. In this paper, the Huber kernel function is used for the subsequent related experiments. The integrated mean square group metric ${c'_{i,j}}$ and the fixed mean square group metric ${c'_j}$ are defined as

\begin{equation}
{c'_{i,j}} = {{{c_{i,j}}} \mathord{\left/
 {\vphantom {{{c_{i,j}}} {{N_{i,j}}}}} \right.
 \kern-\nulldelimiterspace} {{N_{i,j}}}} 
\end{equation}
\begin{equation}
{{{{c'}_j} = {c_j}} \mathord{\left/
 {\vphantom {{{{c'}_j} = {c_j}} {N_j^f}}} \right.
 \kern-\nulldelimiterspace} {N_j^f}}
\end{equation}

Essentially, ${c'_{i,j}}$ calculates the mean square of all measurement point-to-plane errors generated by the plane landmark ${{\bf{\pi }}_j}$ at keyframe ${\mathbb{F}_i}$. Similarly, ${c'_{i,j}}$ is the mean square calculation of all fixed point-to-plane errors for the plane landmark ${{\bf{\pi }}_j}$. Next, we will derive an explicit second-order estimator to solve (7).	

\subsection{Explicit Second-order Estimator}

Using the integrated mean square group metric $\rho \left( {{{c'}_{i,j}}} \right)$ and fixed mean square group metric $\rho \left( {{{c'}_j}} \right)$ with robust kernel functions as examples, this section will derive the Hessian matrix and gradient vector required for solving optimization problem. The proposed solution is regarded as an explicit second-order estimator. Initially, we ignore the effect of $\rho \left(  \cdot  \right)$ and parameterize the estimated values of the mean square group metric ${c'_{i,j}}$ as ${{\bf{x}}_{i,j}} = {\left[ {{t_{i,x}},{t_{i,y}},{t_{i,z}},{\varphi _{i,x}},{\varphi _{i,y}},{\varphi _{i,z}},{\Pi _{j,1}},{\Pi _{j,2}},{\Pi _{j,3}}} \right]^T} = {\left[ {{x_1},{x_2},{x_3},{x_4},{x_5},{x_6},{x_7},{x_8},{x_9}} \right]^T}$. For a given pose ${{\bf{T}}_i} = \left[ \begin{array}{l}
{{\bf{R}}_i}\;\;{{\bf{t}}_i}\\
\;0\;\;\;1
\end{array} \right] \in {\rm{SE(3)}}$, its rotational component ${{\bf{R}}_i} \in {\rm{SO}}(3)$ can be represented using Euler angles

\begin{equation}
\small
{{\bf{R}}_i} = {{\bf{R}}_x}{{\bf{R}}_y}{{\bf{R}}_z} = \left[ \begin{array}{l}
\;\;\;\;\;{c_y}{c_z}\;\;\;\;\;\;\;\;\;\;\;\; - {c_y}{s_z}\;\;\;\;\;\;\;\;\;\;\;\;\;\;\;\;\;\;{s_y}\\
{c_x}{s_z} + {s_x}{s_y}{c_z}\;\;\;\;{c_x}{c_z} - {s_x}{s_y}{s_z}\;\; - {s_x}{c_y}\\
{s_x}{s_z} - {c_x}{s_y}{c_z}\;\;\;\;{c_x}{s_y}{s_z} + {s_x}{c_z}\;\;\;\;{c_x}{c_y}
\end{array} \right].
\end{equation}

\noindent where, 
\begin{equation}
\begin{array}{l}
{s_x} = \sin ({\varphi _{i,x}}),{s_y} = \sin ({\varphi _{i,y}}),{s_z} = \sin ({\varphi _{i,z}}),\\
{c_x} = \cos ({\varphi _{i,x}}),{c_y} = \cos ({\varphi _{i,y}}),{c_z} = \cos ({\varphi _{i,z}}).
\end{array}
\end{equation}

\noindent ${{\bf{t}}_i} \in {\mathbb{R}^3}$ is the translation component of ${{\bf{T}}_i}$. ${{\bf{\Pi }}_j} = {\left[ {{\Pi _{j,1}},{\Pi _{j,2}},{\Pi _{j,3}}} \right]^T}$ is the closest point representation \cite{geneva2018lips} of plane landmark ${{\bf{\pi }}_j}$. Let ${{\bf{\pi '}}_{i,j}} = {\bf{T}}_i^T{{\bf{\pi }}_j}$, ${c'_{i,j}}$ can be rephrased as

\begin{equation}
{c'_{i,j}} = {{{\bf{\pi '}}_{i,j}^T{{\bf{Q}}_{i,j}}{{{\bf{\pi '}}}_{i,j}}} \mathord{\left/
 {\vphantom {{{\bf{\pi '}}_{i,j}^T{{\bf{Q}}_{i,j}}{{{\bf{\pi '}}}_{i,j}}} {{N_{i,j}}}}} \right.
 \kern-\nulldelimiterspace} {{N_{i,j}}}}
\end{equation}

The gradient vector ${{\bf{g}}_{i,j}} = \left[ { \cdots \;\;g_{i,j}^k\; \cdots } \right]_{9 \times 1}^T$ can be obtained by calculating the first-order partial derivative of ${c'_{i,j}}$ with respect to ${{\bf{x}}_{i,j}}$

\begin{equation}
g_{i,j}^k = \frac{{\partial {{c'}_{i,j}}}}{{\partial {x_k}}} = \frac{2}{{{N_{i,j}}}}{\bf{\pi '}}_{i,j}^T{{\bf{Q}}_{i,j}}\frac{{\partial {{{\bf{\pi '}}}_{i,j}}}}{{\partial {x_k}}}
\end{equation}

Hessian matrix ${{\bf{H}}_{i,j}} = {\left[ \begin{array}{l}
\; \ddots \;\;\;\; \vdots \;\;\;\;\; \vdots \;\;\;\\
 \ldots \;\;H_{i,j}^{k,l}\;\;...\\
\;\; \vdots \;\;\;\;\;\; \vdots \;\;\;\; \ddots \;
\end{array} \right]_{9 \times 9}}$ is obtained by taking the second-order partial derivative of ${c'_{i,j}}$ with respect to ${{\bf{x}}_{i,j}}$

\begin{equation}
\small
H_{i,j}^{k,l} = \frac{{{\partial ^2}{{c'}_{i,j}}}}{{\partial {x_k}\partial {x_l}}} = \frac{2}{{{N_{i,j}}}}\left( {{{\frac{{\partial {{{\bf{\pi '}}}_{i,j}}}}{{\partial {x_l}}}}^T}{{\bf{Q}}_{i,j}}\frac{{\partial {{{\bf{\pi '}}}_{i,j}}}}{{\partial {x_k}}} + {\bf{\pi '}}_{i,j}^T{{\bf{Q}}_{i,j}}\frac{{{\partial ^2}{{{\bf{\pi '}}}_{i,j}}}}{{\partial {x_k}\partial {x_l}}}} \right)
\end{equation}

Substituting ${{\bf{R}}_i}$ and ${{\bf{t}}_i}$into ${{\bf{\pi'}}_{i,j}}$, we obtain

\begin{equation}
\small
\begin{array}{l}
{{{\bf{\pi '}}}_{i,j}} = {\bf{T}}_i^T{{\bf{\pi }}_j} = \left[ \begin{array}{l}
{\bf{R}}_i^T\;\;0\\
\;{\bf{t}}_i^T\;\;\;1
\end{array} \right]\left[ \begin{array}{l}
{{\bf{n}}_j}\\
{d_j}
\end{array} \right] = \left[ \begin{array}{l}
\;\;\;{\bf{R}}_i^T{{\bf{n}}_j}\\
{\bf{t}}_i^T{{\bf{n}}_j} + {d_j}
\end{array} \right] = \\
\left[ \begin{array}{l}
{n_{j,x}}({c_y}{c_z}) + {n_{j,y}}({c_x}{s_z} + {s_x}{s_y}{c_z}) + {n_{j,z}}({s_x}{s_z} - {c_x}{s_y}{c_z})\\
{n_{j,x}}( - {c_y}{s_z}) + {n_{j,y}}({c_x}{c_z} - {s_x}{s_y}{s_z}) + {n_{j,z}}({c_x}{s_y}{s_z} + {s_x}{c_z})\\
{n_{j,x}}({s_y}) + {n_{j,y}}( - {s_x}{c_y}) + {n_{j,z}}({c_x}{c_y})\\
{t_{i,x}}{n_{j,x}} + {t_{i,y}}{n_{j,y}} + {t_{i,z}}{n_{j,z}} + {d_j}
\end{array} \right]
\end{array}
\end{equation}

Then, by calculating the first-order partial derivative of ${{\bf{\pi '}}_{i,j}}$ with respect to ${{\bf{x}}_{i,j}}$, ${\bf{J}}_{i,j}^k = \frac{{\partial {{{\bf{\pi '}}}_{i,j}}}}{{\partial {x_k}}}$ can be obtained. We record the calculation results ${{\bf{J}}_{i,j}}$ as

\begin{equation}
{{\bf{J}}_{i,j}} = {\left[ {\frac{{\partial {{{\bf{\pi '}}}_{i,j}}}}{{\partial {{\bf{t}}_i}}},\frac{{\partial {{{\bf{\pi '}}}_{i,j}}}}{{\partial {{\bf{\varphi }}_i}}},\frac{{\partial {{{\bf{\pi '}}}_{i,j}}}}{{\partial {{\bf{\Pi }}_j}}}} \right]^T}.
\end{equation}

\noindent where,

\begin{equation}
\small
\frac{{\partial {{{\bf{\pi '}}}_{i,j}}}}{{\partial {{\bf{t}}_i}}} = \left[ {\frac{{\partial {{{\bf{\pi '}}}_{i,j}}}}{{\partial {x_1}}},\frac{{\partial {{{\bf{\pi '}}}_{i,j}}}}{{\partial {x_2}}},\frac{{\partial {{{\bf{\pi '}}}_{i,j}}}}{{\partial {x_3}}}} \right] = \left[ \begin{array}{l}
0\;\;\;\;\;0\;\;\;\;\;0\\
0\;\;\;\;\;0\;\;\;\;\;0\\
0\;\;\;\;\;0\;\;\;\;\;0\\
{n_{j,x}}\;\;{n_{j,y}}\;\;{n_{j,z}}
\end{array} \right]
\end{equation}

\begin{equation}
\small
\frac{{\partial {{{\bf{\pi '}}}_{i,j}}}}{{\partial {{\bf{\varphi }}_i}}} = \left[ {\frac{{\partial {{{\bf{\pi '}}}_{i,j}}}}{{\partial {x_4}}},\frac{{\partial {{{\bf{\pi '}}}_{i,j}}}}{{\partial {x_5}}},\frac{{\partial {{{\bf{\pi '}}}_{i,j}}}}{{\partial {x_6}}}} \right] = \left[ \begin{array}{l}
a\;\;\;d\;\;g\\
b\;\;\;e\;\;\;h\\
c\;\;\;f\;\;0\\
0\;\;\;0\;\;\;0
\end{array} \right]
\end{equation}

\begin{equation}
\small
\frac{{\partial {{{\bf{\pi '}}}_{i,j}}}}{{\partial {{\bf{\Pi }}_j}}} = \frac{{\partial {{{\bf{\pi '}}}_{i,j}}}}{{\partial {{\bf{\pi }}_j}}}\frac{{\partial {{\bf{\pi }}_j}}}{{\partial {{\bf{\Pi }}_j}}} = {\bf{T}}_i^T\left[ \begin{array}{l}
\frac{1}{{{d_j}}}({{\bf{I}}_{3 \times 3}} - {{\bf{n}}_j}{{\bf{n}}_j}^T)\\
\;\;\;\;\;\;\;\;\;{{\bf{n}}_j}^T
\end{array} \right]
\end{equation}

For simplification reason, the full expression of (18) is show in Supplementary Materials. We define ${\bf{\vec G}}_{i,j}^{k,l} = \frac{{{\partial ^2}{{{\bf{\pi '}}}_{i,j}}}}{{\partial {x_k}\partial {x_l}}}$ as

\begin{equation}
\begin{array}{l}
{{\bf{G}}_{i,j}} = {\left[ \begin{array}{l}
{\bf{\vec G}}_{i,j}^{1,1}\; \cdots \;\;{\bf{\vec G}}_{i,j}^{1,9}\\
\; \vdots \;\;\;\;\; \ddots \;\;\; \vdots \\
{\bf{\vec G}}_{i,j}^{9,1}\; \cdots \;\;{\bf{\vec G}}_{i,j}^{9,9}\;
\end{array} \right]_{36 \times 9}} = \\
\left[ \begin{array}{l}
\vec 0\;\;\;\;\vec 0\;\;\;\;\vec 0\;\;\;\;\;\vec 0\;\;\;\;\;\vec 0\;\;\;\;\;\vec 0\;\;\;\;\vec a\;\;\;\;\vec b\;\;\;\;\vec c\\
\vec 0\;\;\;\;\vec 0\;\;\;\;\vec 0\;\;\;\;\;\vec 0\;\;\;\;\;\vec 0\;\;\;\;\;\vec 0\;\;\;\;\vec d\;\;\;\;\vec e\;\;\;\vec f\\
\vec 0\;\;\;\;\vec 0\;\;\;\;\vec 0\;\;\;\;\;\vec 0\;\;\;\;\;\vec 0\;\;\;\;\;\vec 0\;\;\;\;\vec g\;\;\;\;\vec h\;\;\;\;\vec i\\
\vec 0\;\;\;\;\vec 0\;\;\;\;\vec 0\;\;\;\;\;\vec j\;\;\;\;\vec k\;\;\;\;\;\vec l\;\;\;\;\vec p\;\;\;\;\vec s\;\;\;\;\vec v\\
\vec 0\;\;\;\;\vec 0\;\;\;\;\vec 0\;\;\;\;\;\vec k\;\;\;\;\vec m\;\;\;\;\vec n\;\;\;\;\vec q\;\;\;\;\vec t\;\;\;\;\vec w\\
\vec 0\;\;\;\;\vec 0\;\;\;\;\vec 0\;\;\;\;\;\vec l\;\;\;\;\vec n\;\;\;\;\vec o\;\;\;\;\;\vec r\;\;\;\;\vec u\;\;\;\;\vec x\\
\vec a\;\;\;\;\vec d\;\;\;\;\vec g\;\;\;\;\vec p\;\;\;\;\vec q\;\;\;\;\vec r\;\;\;\;\vec y\;\;\;\;\vec z\;\;\;\;\vec \alpha \\
\vec b\;\;\;\;\vec e\;\;\;\;\vec h\;\;\;\;\vec s\;\;\;\;\;\vec t\;\;\;\;\vec u\;\;\;\;\vec z\;\;\;\;\vec \beta \;\;\;\;\vec \gamma \\
\vec c\;\;\;\;\vec f\;\;\;\;\vec i\;\;\;\;\vec v\;\;\;\;\vec w\;\;\;\;\vec x\;\;\;\;\vec \alpha \;\;\;\;\vec \gamma \;\;\;\;\vec \eta 
\end{array} \right]
\end{array}
\end{equation}

The detailed calculation of (20) is in Supplementary Materials. A similar derivation method is used for the fixed mean square group metric. The estimated values of the fixed mean square group metric denote the landmarks ${{\bf{\pi }}_j}$ on the plane, which can be parameterized as ${{\bf{y}}_j} = {\left[ {{\Pi _{j,1}},{\Pi _{j,2}},{\Pi _{j,3}}} \right]^T} = {\left[ {{y_1},{y_2},{y_3}} \right]^T}$. The gradient vector is derived by calculating the first-order partial derivative of ${c'_j}$ with respect to ${{\bf{y}}_j}$

\begin{equation}
g_j^k = \frac{{\partial {{c'}_j}}}{{\partial {y_k}}} = \frac{2}{{N_j^f}}{\bf{\pi }}_j^T{{\bf{M}}_j}\frac{{\partial {{\bf{\pi }}_j}}}{{\partial {y_k}}}
\end{equation}

Hessian matrix ${{\bf{H}}_j} = {\left[ \begin{array}{l}
\; \ddots \;\;\;\; \vdots \;\;\;\;\; \vdots \;\;\;\\
 \ldots \;\;H_j^{k,l}\;\;...\\
\;\; \vdots \;\;\;\;\;\;\; \vdots \;\;\; \ddots \;
\end{array} \right]_{3 \times 3}}$ represents the second-order partial derivative of ${c'_j}$ with respect to ${{\bf{y}}_j}$ 

\begin{equation}
H_j^{k,l} = \frac{{{\partial ^2}{{c'}_j}}}{{\partial {y_k}\partial {y_l}}} = \frac{2}{{N_j^f}}\left( {{{\frac{{\partial {{\bf{\pi }}_j}}}{{\partial {y_l}}}}^T}{{\bf{M}}_j}\frac{{\partial {{\bf{\pi }}_j}}}{{\partial {y_k}}} + {\bf{\pi }}_j^T{{\bf{M}}_j}\frac{{{\partial ^2}{{\bf{\pi }}_j}}}{{\partial {y_k}\partial {y_l}}}} \right)
\end{equation}

By calculating the first-order partial derivative of ${{\bf{\pi }}_j}$ with respect to ${{\bf{y}}_j}$, ${\bf{J}}_j^k = \frac{{\partial {{\bf{\pi }}_j}}}{{\partial {y_k}}}$ can be obtained

\begin{equation}
{{\bf{J}}_j} = \left[ {\frac{{\partial {{\bf{\pi }}_j}}}{{\partial {y_1}}},\frac{{\partial {{\bf{\pi }}_j}}}{{\partial {y_2}}},\frac{{\partial {{\bf{\pi }}_j}}}{{\partial {y_3}}}} \right] = \frac{{\partial {{\bf{\pi }}_j}}}{{\partial {{\bf{\Pi }}_j}}}.
\end{equation}

Then, ${\bf{\vec G}}_j^{k,l} = \frac{{{\partial ^2}{{\bf{\pi }}_j}}}{{\partial {y_k}\partial {y_l}}}$ can be defined as

\begin{equation}
{{\bf{G}}_j} = \left[ \begin{array}{l}
{\bf{\vec G}}_j^{1,1}\;\;{\bf{\vec G}}_j^{1,2}\;\;{\bf{\vec G}}_j^{1,3}\\
{\bf{\vec G}}_j^{2,1}\;\;{\bf{\vec G}}_j^{2,2}\;\;{\bf{\vec G}}_j^{2,3}\\
{\bf{\vec G}}_j^{3,1}\;\;{\bf{\vec G}}_j^{3,2}\;\;{\bf{\vec G}}_j^{3,3}\;
\end{array} \right] = \left[ \begin{array}{l}
{{\vec a}^f}\;\;\;\;{{\vec b}^f}\;\;\;\;{{\vec c}^f}\\
{{\vec b}^f}\;\;\;\;{{\vec d}^f}\;\;\;\;{{\vec e}^f}\\
{{\vec c}^f}\;\;\;\;{{\vec e}^f}\;\;\;\;{{\vec f}^f}
\end{array} \right]
\end{equation}

For a detailed calculation of (24), please refer to the Supplementary Materials. Finally, we need to derive the Hessian matrix and gradient vector considering $\rho \left(  \cdot  \right)$. We will use $c$ in the following representation since ${c'_{i,j}}$ and ${c'_j}$ share the same processing methods. The estimated values are defined as ${\bf{x}}$. By performing Taylor expansion on $\rho \left( c \right)$ and retaining terms up to the second order, we obtain	

\begin{equation}
\rho \left( c \right) = \text{const} + \dot{\rho} \Delta c + \frac{1}{2}\ddot{\rho}\Delta {c^2}
\end{equation}

\noindent where, $\text{const}$ represents the constant. $\dot{\rho}$ and $\ddot{\rho}$ denote the first and second derivatives of $\rho \left( c \right)$ with respect to the metric $c$, respectively. The incremental metric $\Delta c$ has the following form

\begin{equation}
\Delta c = c\left( {{\bf{x}} + \Delta {\bf{x}}} \right) - c\left( {\bf{x}} \right) = {\bf{J}}\Delta {\bf{x}} + \frac{1}{2}\Delta {{\bf{x}}^T}{\bf{H}}\Delta {\bf{x}}
\end{equation}

Substituting (26) into (25), we obtain

\begin{equation}
\begin{array}{l}
\rho \left( c \right) = const + \dot{\rho}\Delta c + \frac{1}{2}\ddot{\rho}\Delta {c^2}\\
\;\;\;\;\;\;\; \approx const + \dot{\rho}{\bf{J}}\Delta {\bf{x}} + \frac{1}{2}\dot{\rho}\Delta {{\bf{x}}^T}{\bf{H}}\Delta {\bf{x}} + \frac{1}{2}\ddot{\rho}\Delta {{\bf{x}}^T}{{\bf{J}}^T}{\bf{J}}\Delta {\bf{x}}\\
\;\;\;\;\;\;\; = const + \dot{\rho}{\bf{J}}\Delta {\bf{x}} + \frac{1}{2}\Delta {{\bf{x}}^T}\left( {\dot{\rho}{\bf{H}} + \ddot{\rho}{{\bf{J}}^T}{\bf{J}}} \right)\Delta {\bf{x}}
\end{array}
\end{equation}

By taking the derivative of $\Delta {\bf{x}}$ in (27) and making it equal to zero, the normal equation considering the robust kernel function can be obtained

\begin{equation}
\left( {\dot{\rho}{\bf{H}} + \ddot{\rho}{{\bf{J}}^T}{\bf{J}}} \right)\Delta {\bf{x}} =  - \dot{\rho}{{\bf{J}}^T}
\end{equation}

Applying normal equation calculations on each metric in (7), we can obtain an estimation method which is similar to solving the visual BA-based optimization problem. Then, the Schur complement technique \cite{triggs2000bundle} is employed to accelerate the solution.

\section{Integrated System}

This section will present a loosely coupled SLAM system that integrates the proposed RSO-BA estimator. The system consists of two parts: front-end and back-end. The front end can calculate the odometry pose and undistorted point cloud of the current frame. If the current frame is identified as a keyframe, it will be inserted into the system's back end. The RSO-BA estimator is then used to refine the map.

\subsection{Front End}

Assuming the odometry coordinate system is $o$. The system's front end uses the Fast-LIO2 \cite{xu2022fast} module to process the raw measurement point cloud and IMU data. A local point cloud map is maintained using the ikd-tree \cite{cai2021ikd} data structure. The distortion in the raw point cloud ${{\bf{\psi }}_i}$ is corrected using a bidirectional propagation algorithm based on the IMU sensor. The iterated Kalman filter (IEKF) \cite{bell1993iterated} estimator is then used to register the undistorted point cloud ${{\bf{\bar \psi }}_i}$ to the local point cloud map. The estimated odometry pose for the current LiDAR frame is ${{\bf{T}}_{ol,i}}$.

Then, keyframe is created if the following two conditions are satisfied

(1) The difference in timestamps between the current frame and the last keyframe exceeds the threshold $t{h_{time}}$. 

(2) The 2-norm of the translation increment or the 2-norm of the rotation increment, compared to the last keyframe, is greater than $t{h_{pos}}$ or $t{h_{\deg }}$, respectively.

\noindent The undistorted point cloud ${{\bf{\bar \psi }}_i}$ and estimated odometry pose ${{\bf{T}}_{ol,i}}$ are used to create the current keyframe ${\mathbb{F}_k}$. The keyframe ${\mathbb{F}_k}$ is then inserted into the system's backend.	

\subsection{Back End}

The system's backend employs the hash adaptive voxel map (H-AV) proposed by BALM \cite{liu2021balm} for global map maintenance. Similar to LOAM \cite{zhang2014loam}, we use a corrected transformation ${{\bf{T}}_{mo}}$ from odometry coordinate system $o$ to map coordinate system $m$. Based on the odometry pose ${{\bf{T}}_{ol}}$ calculated by the front end and the latest corrected transformation ${{\bf{T}}_{mo}}$, the predicted pose ${{\bf{T'}}_{ml}}$ in the map coordinate system is defined as

\begin{equation}
{{\bf{T'}}_{ml}} = {{\bf{T}}_{mo}}{{\bf{T}}_{ol}}
\end{equation}

The 9-value sampling algorithm \cite{deschaud2018imls} is used to extract a fixed number of plane feature point set ${{\bf{\bar \Phi }}_k}$ with complete pose constraints from ${{\bf{\bar \psi }}_k}$. The keyframe ${\mathbb{F}_k}$ storing ${{\bf{\bar \psi }}_k}$ and ${{\bf{T'}}_{ml}}$ is then inserted into the sliding window. Then, we use ${{\bf{T'}}_{ml}}$ to insert ${{\bf{\bar \psi }}_k}$ into the H-AV map and perform adaptive voxel segmentation. When the number of keyframes in the sliding window reaches $t{h_{kf}}$, the RSO-BA estimator proposed in Section III.C works for map refinement. ${{\bf{T}}_{mo}}$ is corrected by the optimized keyframe pose ${{\bf{T}}_{ml}}$ and odometry pose ${{\bf{T}}_{ol}}$ 

\begin{equation}
{{\bf{T}}_{mo}} = {{\bf{T}}_{ml}}{\bf{T}}_{ol}^{ - 1}
\end{equation}

Finally, the measurement information corresponding to each landmark in the oldest keyframe is marginalized to the fixed group matrix, after which the oldest keyframe is removed from the sliding window.

\section{Experiment}

This section will verify the effectiveness of the proposed RSO-BA algorithm. All experiments were conducted on the AMD ® Ryzen 7 5800h (8 cores @ 3.2 GHz), 16GB RAM, ROS Melodic environment. The experiment is divided into two parts: structured scene experiments and unstructured scene experiments.

\subsection{Experimental Setup}

We compare the proposed methods with many different algorithms which are regarded as the state-of-the-art methods. For the other BA algorithms involved in the experiment, BALM2 \cite{liu2023efficient} is an improved version based on BALM \cite{liu2021balm}. BALM2 introduces a two-step BA algorithm that iteratively performs eigenvalue fitting and multi-view pose registration. The multi-view pose registration is achieved using a second-order Newton estimator. By introducing the concept of "point clustering", the Hessian matrix and gradient vector can be calculated as a "group", effectively avoiding the time-consuming process of point-by-point calculation. Unlike BALM2, PA \cite{zhou2021lidar} estimates plane landmarks and poses simultaneously. This algorithm uses QR decomposition to calculate the reduction error vector and the reduction Jacobian matrix. Then, an approximate second-order algorithm based on levenberg-marquardt (LM) \cite{ranganathan2004levenberg} is used for updating state variables.

In the structured scene experiment, we validated the proposed algorithm using the SLAM system developed in Section IV. For the indoor Hilti \cite{helmberger2022hilti} dataset, the measurement point cloud resolution ${r^{mea}}$ and ikd-tree map resolution ${r^{ikd}}$ in Fast-LIO2 \cite{xu2022fast} are set to 0.4 m and 0.2 m, respectively. The keyframe detection conditions are $t{h_{time}} = 0.25s$, $t{h_{pos}} = 0.1m$, and $t{h_{\deg }} = {0.05^ \circ }$. The maximum depth and maximum voxel resolution of the Hash Adaptive Voxel (H-AV) map are ${d_{\max }} = 4$ and $r_{\max }^{hav} = 1m$. The number of keyframes in the sliding window is set to $t{h_{kf}} = 10$. For the outdoor UrbanNav \cite{hsu2021urbannav} dataset, the measurement point cloud resolution ${r^{mea}}$ and map resolution ${r^{ikd}}$ are both set to 0.5 m. The keyframe detection remains the same as above. The maximum depth of the H-AV map is ${d_{\max }} = 3$, and the maximum voxel resolution is $r_{\max }^{hav} = 2m$. The number of keyframes in the sliding window is consistent with the indoor experiment.

We also compared the performance of different BA algorithms in unstructured scenes using the ETHZ \cite{pomerleau2012challenging} dataset. The initial pose estimation method for the LiDAR frame is detailed in Section V.C. The back end parameter settings are the same as those on the Hilit  \cite{helmberger2022hilti} dataset. To ensure a fair comparison, all BA estimators in the experiment are iteratively optimized using the LM \cite{ranganathan2004levenberg} algorithm. The damping factor $\lambda$ for all algorithms is set to 0.01 with a maximum of 20 iterations. For the indoor Hilti and ETHZ point cloud registration datasets, the Huber kernel function threshold of the RSO-BA algorithm is set to 0.02. For the outdoor UrbanNav dataset, the threshold is set to 0.1.

\begin{table}[b]
\caption{Registration accuracy under the Hilti dataset.}
\centering
\begin{tabular}{c|c|c|c}
\toprule  
\multicolumn{4}{c}{{Absolute Motion Trajectory RMSE [cm]}}\\
\toprule  
\bf{Sequence} & \bf{BALM2} & \bf{PA} & \bf{RSO-BA} \\
\specialrule{0em}{0pt}{2pt}
\toprule  
LAB\_2 & 1.4 & 2.5 & \bf{1.3} \\
uzh\_run2 & \bf{16.8} & 25.4 & 16.9  \\
exp04\_level & 2.1 & \bf{2.0} & \bf{2.0}  \\
exp05\_level\_2 & \bf{-} & \bf{2.1} & \bf{2.1} \\
exp06\_level\_3 & \bf{-} & 3.9 & \bf{2.7}  \\
exp14\_basement\_2 & \bf{-} & 6.6 & \bf{3.5}  \\
\bottomrule 
\specialrule{0em}{0pt}{2pt}
Mean & \bf{-} & 7.1 & 4.8 \\
\bottomrule 
\end{tabular}
\caption*{Note: '{\bf{-}}' means the sequence was not successfully run entirely. Bold values represent the optimal registration accuracy.}
\end{table}

\subsection{Structured Scene Experiment}

In this section, we conduct experiments using the indoor Hilti \cite{helmberger2022hilti} dataset and the outdoor UrbanNav \cite{hsu2021urbannav} dataset to evaluate the registration accuracy of different BA algorithms using the ATE \cite{sturm2012benchmark} metric. The Hilti dataset includes raw point clouds collected by Ouster Os0-64 or Hesai PandarXT-32 LiDAR, along with acceleration and angular velocity data collected by an Analong Devices ADIS1644 inertial measurement unit. Data collection was conducted in static scenarios such as laboratories and offices. The UrbanNav dataset provides raw point clouds collected by a Velodyne HDL-32E LiDAR, along with acceleration and angular velocity data from an Xsens Mti 10 inertial measurement unit. The dataset was collected in Hong Kong and involves numerous moving vehicles, which can easily disrupt the stability of the system's front end. Figures 2 show the robot trajectories of different BA algorithms on the UrbanNav dataset.	

\begin{figure}[t]
\centering
\includegraphics[scale=0.92]{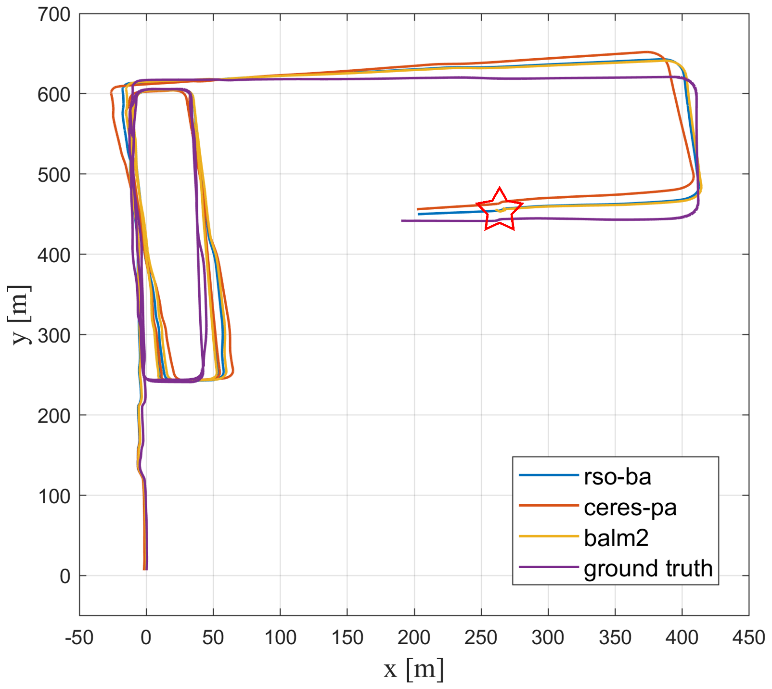}
\caption{The running trajectories of different BA algorithms under UrbanNav Mongkok sequence. The accuracy of PA is much lower than that of RSO-BA and BALM2. BALM2 experienced severe trajectory drift at the location marked by the red star..}
\label{fig_framework5}
\end{figure}

\begin{figure}[h]
\centering
\vspace{-0.5cm} 
\begin{subfigure}[b]{0.28\textwidth}
\centering
\includegraphics[width=\textwidth]{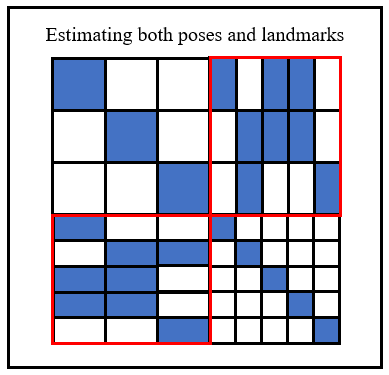}
\caption{Explicit BA}
\end{subfigure}%
\hfill
\begin{subfigure}[b]{0.42\textwidth}
\centering
\includegraphics[width=\textwidth]{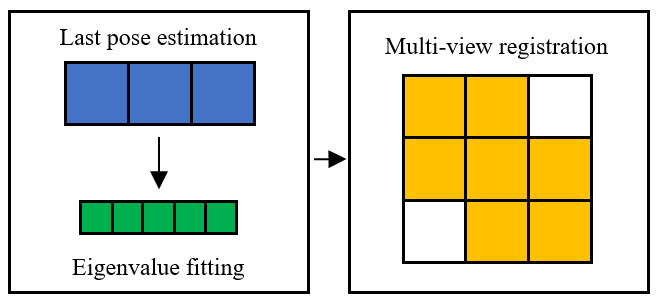}
\caption{Implicit BA}
\end{subfigure}%
\caption{Single iteration of different BA algorithms.}
\label{fig:t-SNE1.0}
\end{figure}

Tables I and II demonstrate that the proposed RSO-BA algorithm achieves the best ATE accuracy in most sequences. Among all successful sequences, it is evident that the estimation accuracy of RSO-BA and BALM2 is much better than that of PA. The primary reason for the superior performance of RSO-BA and BALM2 is that both estimators utilize the second-order Newton's method, which leverages second-order terms to achieve better convergence performance. Additionally, compared to BALM2, RSO-BA offers certain advantages in registration accuracy. This is because, in BALM2’s two-step estimation process, eigenvalue fitting (landmark estimation) only depends on the multi-view pose and measurement information from the previous step. The proposed approach overlooks the interaction between pose and landmarks during the current state update process (highlighted in the red box in Figure 3(a)), an interaction that is explicitly accounted for in the RSO-BA algorithm.

\begin{table}[h]
\caption{Registration accuracy under the UrbanNav dataset. }
\centering
\begin{tabular}{c|c|c|c}
\toprule  
\multicolumn{4}{c}{{Absolute Motion Trajectory RMSE [m]}}\\
\toprule  
\bf{Sequence} & \bf{BALM2} & \bf{PA} & \bf{RSO-BA} \\
\specialrule{0em}{0pt}{2pt}
\toprule  
HK-Mongkok & 7.11 & 12.77 & \bf{6.77} \\
HK-TST & \bf{-} & 12.1 & \bf{8.97} \\
HK-Whampoa & \bf{-} & 9.89 & \bf{6.83}  \\
\bottomrule 
\specialrule{0em}{0pt}{2pt}
Mean & \bf{-} & 11.58 & \bf{7.52} \\
\bottomrule 
\end{tabular}
\caption*{Note: '{\bf{-}}' means the sequence was not successfully run entirely. Bold values represent the optimal registration accuracy.}
\end{table}

On the other hand, we observed that BALM2 failed to run in multiple sequences. One reason is that the BALM2 algorithm requires the current multi-view pose for landmark estimation at the first step in each iteration. If the initial value provided by the front-end pose is poor, the landmark parameters may not be able to converge to a stable state, leading to divergent estimation results. In contrast, the RSO-BA and PA algorithms, which estimate landmarks and poses simultaneously, can leverage the constraints between them. This makes the estimation model more robust, allowing it to successfully execute across all sequences.

\subsection{Unstructured Scene Experiment}

The ETHZ \cite{pomerleau2012challenging} point cloud registration dataset statically collected measurement point clouds using the Hokuyo UTM-30LX LiDAR. This dataset includes a range of complex scenarios, from structured environments (apartments) to unstructured scenarios (forests). Unstructured scenarios present challenges for feature-based BA schemes. Since the measurement point clouds do not require motion distortion correction, we utilize the point-to-plane error metric proposed by \cite{chen1992object} and the scan-to-map registration method to estimate the initial odometry pose required for the system's backend. The map is maintained incrementally using the ikd-tree data structure. Given that all measurement point clouds are captured from independent perspectives, we directly insert the measurement point clouds and estimated pose into the backend without performing keyframe detection. 

\begin{table}[h]
\caption{Registration accuracy under the ETHZ dataset.}
\centering
\begin{tabular}{c|c|c|c}
\toprule  
\multicolumn{4}{c}{{Absolute Motion Trajectory RMSE [cm]}}\\
\toprule  
\bf{Sequence} & \bf{BALM2} & \bf{PA} & \bf{RSO-BA} \\
\specialrule{0em}{0pt}{2pt}
\toprule  
apartment & \bf{0.8} & 1.3 & \bf{0.8} \\
hauptgebaude & \bf{0.4} & 0.6 & 0.5  \\
gazebo\_summer & 1.7 & 11.2 & \bf{1.4} \\
gazebo\_winter & 8.6 & 3.1 & \bf{1.1} \\
mountain\_plain & 12.1 & \bf{6.1} & 6.9  \\
stairs & 1.2 & 1.1 & \bf{0.9}  \\
wood\_autumn & 2.0 & 2.4 & \bf{1.5}  \\
wood\_summer & 1.8 & 3.1 & \bf{1.7}  \\
\bottomrule 
\specialrule{0em}{0pt}{2pt}
Mean & 3.5 & 3.6 & \bf{1.8} \\
\bottomrule 
\end{tabular}
\caption*{Note: Bold values represent the optimal registration accuracy.}
\end{table}

Table III presents the ATE evaluation results for three BA algorithms. In structured scenes such as apartments and stairs, the registration accuracy of the proposed RSO-BA and BALM2 algorithms is similar and obviously better than that of the PA algorithm, consistent with the findings in Section V.B. However, in unstructured scenes such as mountain and woods, the RSO-BA algorithm significantly outperforms both BALM2 and PA. This improvement is due to the RSO-BA algorithm’s robust kernel function which reweights all measurements. The measurements corresponding to flatter plane landmarks have the larger weights in the optimization process, that improves estimation accuracy. 

\begin{table}[h]
\caption{Map quality evaluation under the ETHZ dataset.}
\centering
\begin{tabular}{c|c|c|c}
\toprule  
\multicolumn{4}{c}{{Voxel occupancy number}}\\
\toprule  
\bf{Sequence} & \bf{BALM2} & \bf{PA} & \bf{RSO-BA} \\
\specialrule{0em}{0pt}{2pt}
\toprule  
apartment & \bf{56910} & 57664 & 57011 \\
hauptgebaude & 225853 & 225812 & \bf{225671}  \\
gazebo\_summer & 155449 & 182999 & \bf{154424} \\
gazebo\_winter & 195433 & 197637 & \bf{177688} \\
mountain\_plain & 92149 & \bf{90087} & 90744 \\
stairs & 88501 & 88675 & \bf{88169}  \\
wood\_autumn & 349279 & 346779 & \bf{346219}  \\
wood\_summer & 411731 & 410541 & \bf{409234}   \\
\bottomrule 
\specialrule{0em}{0pt}{2pt}
Mean & 196913 & 200024 & \bf{193645} \\
\bottomrule 
\end{tabular}
\caption*{Note: Bold values represent the minimum voxel occupancy.}
\end{table}

Table IV presents the quality evaluation of the reconstructed maps using the voxel occupancy \cite{filatov20172d} method. During the experiment, we set the voxel size to 0.1m. A high map reconstruction quality and an accurate sensor pose estimation result will make point cloud map occupy few voxels. The average voxel occupancy number of the RSO-BA algorithm is significantly lower than that of the other two BA algorithms. Additionally, the map quality evaluation results in Table IV are consistent with the registration accuracy evaluation results shown in Table III using different methods.

\section{Conclusions and Future Work}

In this letter, we propose a novel LiDAR bundle adjustment algorithm based on a new designed mean square group metric. Leveraging the mean square group metric with scale interpretability, the robust BA model can be built using a robust kernel function. We then derive the analytical Hessian matrix and gradient vector required for the second-order estimator (RSO-BA) to promote estimation accuracy. The proposed RSO-BA estimator is compared with other advanced LiDAR BA algorithms, such as BALM2 and PA, on publicly available datasets. Experimental results demonstrate that the RSO-BA estimator offers superior robustness and estimation accuracy in complex scenarios, including unstructured and highly dynamic environments. However, the pure LiDAR BA algorithm tends to degrade in weakly textured scenes, such as long corridors, leading to significant trajectory drift. As a high-frequency, internal sensing sensor, the inertial measurement unit (IMU) can effectively overcome the above problem. In the future, we plan to explore the multi-sensor fusion BA estimation algorithm to improve the system's adaptability across diverse environments.

\bibliographystyle{IEEEtran}
\bibliography{mlo.bib}

\end{document}